\documentclass[10pt,twocolumn,letterpaper]{article}

\usepackage[pagenumbers]{cvpr} 

\usepackage{graphicx}
\usepackage{amsmath}
\usepackage{amssymb}
\usepackage{booktabs}

\usepackage{algorithmic}
\usepackage{algorithm}
\usepackage{multirow}

\usepackage{lipsum}

\usepackage[pagebackref,breaklinks,colorlinks]{hyperref}

\usepackage[capitalize]{cleveref}
\usepackage{mathtools}

\crefname{section}{Sec.}{Secs.}
\Crefname{section}{Section}{Sections}
\Crefname{table}{Table}{Tables}
\crefname{table}{Tab.}{Tabs.}

\def\Vec#1{{\boldsymbol{#1}}}
\def\Mat#1{{\boldsymbol{#1}}}

\newcommand{\blog}{{ \rm Log} }
\newcommand{\bexp}{{ \rm Exp} }

\usepackage{xcolor}
\usepackage[normalem]{ulem}

\usepackage[capitalize]{cleveref}
\crefname{section}{Sec.}{Secs.}
\Crefname{section}{Section}{Sections}
\Crefname{table}{Table}{Tables}
\crefname{table}{Tab.}{Tabs.}

\def\cvprPaperID{7524} 
\def\confName{CVPR}
\def\confYear{2023}

\begin{document}

\title{Exploring Data Geometry for Continual Learning}

\author{
Zhi Gao\textsuperscript{\rm 1},
Chen Xu\textsuperscript{\rm 2*},
Feng Li,
Yunde Jia\textsuperscript{\rm 2,1},
Mehrtash Harandi\textsuperscript{\rm 3},
Yuwei Wu\textsuperscript{\rm 1,2*} \\
\textsuperscript{\rm 1}Beijing Key Laboratory of Intelligent Information Technology, \\ School of Computer Science \& Technology, Beijing Institute of Technology, China \\
\textsuperscript{\rm 2}Guangdong Laboratory of Machine Perception and Intelligent Computing, \\
   Shenzhen MSU-BIT University, China \\
\textsuperscript{\rm 3}Department of Electrical and Computer Systems Eng., Monash University, and Data61, Australia \\
{\tt\small \{gaozhi\underline{\hspace{0.5em}}2017,jiayunde,wuyuwei\}@bit.edu.cn} \\
{\tt\small xuchen@smbu.edu.cn, lifeng.passion@gmail.com, mehrtash.harandi@monash.edu}
}

\maketitle

\renewcommand{\thefootnote}{\fnsymbol{footnote}} 
\footnotetext{$^*$ Corresponding authors: Chen Xu and Yuwei Wu.}

\begin{abstract}
Continual learning aims to efficiently learn from a non-stationary stream of data while avoiding forgetting the knowledge of old data. 
In many practical applications, data complies with non-Euclidean geometry. As such, the commonly used Euclidean space cannot gracefully capture non-Euclidean geometric structures of data, leading to inferior results. 
In this paper, we study continual learning from a novel perspective by exploring data geometry for the non-stationary stream of data.
Our method dynamically expands the geometry of the underlying space to match growing geometric structures induced by new data, and prevents forgetting by keeping geometric structures of old data into account. 
In doing so, making use of the mixed curvature space, we propose an incremental search scheme, through which the growing geometric structures are encoded.
Then, we introduce an angular-regularization loss and a neighbor-robustness loss to train the model, capable of penalizing the change of global geometric structures and local geometric structures.
Experiments show that our method achieves better performance than baseline methods designed in Euclidean space.
\end{abstract}

\section{Introduction}
\label{sec:intro}

Unlike humans, artificial neural networks perform poorly to learn new knowledge in a continual manner. The tendency to lose the knowledge previously learned, known as \emph{catastrophic forgetting}, is due to the fact that important parameters of a neural network for old data are changed to meet the objectives of new data.
There have been many continual learning methods~\cite{Zhu2021ClassIncrementalLV,Chaudhry2021UsingHT,Kirkpatrick2017OvercomingCF,Cheraghian2021SemanticawareKD,Mallya2018PackNetAM,Yan2021DERDE}, and their goal is to remember the knowledge from old data while effectively learning from new data.
They have achieved impressive performance in alleviating catastrophic forgetting.

However, a long-lasting issue with existing methods is that data geometry is rarely studied in continual learning.
Existing methods usually assume that data is Euclidean and they use Euclidean geometry to process the data stream.
In fact, data in countless applications intrinsically has non-Euclidean geometric structures~\cite{BronsteinBLSV17,AtighKM21}.
Several studies show that non-Euclidean geometric structures can be better captured by particular forms of Riemannian geometry~\cite{FangHP21,QiYLL21}. 
For example, the hyperbolic geometry has a natural expressive ability for the hierarchical structure and is hence used successfully for fine-grained images~\cite{khrulkov2020hyperbolic,long2020searching}. The spherical geometry is shown as a suitable choice for face images that have the cyclical structure~\cite{Li_2021_CVPR,WangWSWNAXYC21}.
In addition to the geometric structures discussed above, natural data may be diverse and irregular in structure, \eg, 
data exhibits hierarchical forms in some regions and cyclical forms in others~\cite{Shevkunov2021OverlappingSF,lou2021learning}.
Overall, distortions produced when using Euclidean geometry for non-Euclidean geometric structures are overwhelming, causing the loss of semantic information, and hence resulting in inferior performance~\cite{Bachmann2019ConstantCG}.
In this paper, we study how to attain suitable non-Euclidean geometry to capture the intrinsic geometric structures of data during continual learning.

\begin{figure}[!t]
  \centering
  \includegraphics[width=0.48\textwidth]{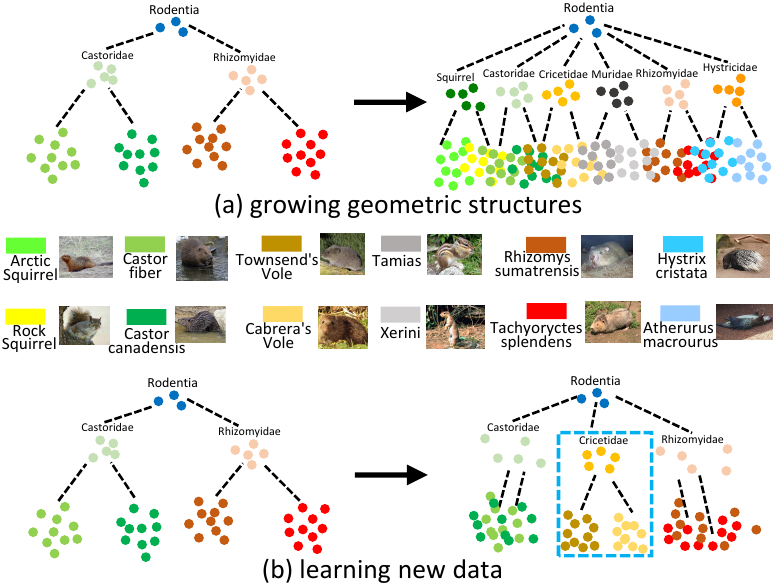}
  \caption{Illustrations of the two challenges of exploring data geometry in continual learning. Different color denotes different classes. (a) A fixed geometry cannot handle more and more complex hierarchy in a data stream. The geometric structure of leaf nodes is destroyed. (b) Learning from new data (in the blue dash box) will destroy the captured hierarchical structures of old data. }
  \label{fig:structure}
\end{figure}

To achieve our goal, we have to face two challenges (see \cref{fig:structure}).
\textbf{(1)} Non-stationary stream of data will inevitably increase the complexity of intrinsic geometric structures.
In other words, fixing the geometry of the underlying space cannot always match new and unseen data in continual learning. For example, more and more complex hierarchies in a data stream bring more leaf nodes,
requiring a faster growing space volume with the radius,
which conflicts with a fixed geometry~\cite{GaoWJH21}.
\textbf{(2)} Old data is not accessible in continual learning, and learning from new data may destroy the captured geometric structures of old data, resulting in the catastrophic forgetting problem. Since geometric structures are characterized by distances or angles between instances, destroying the captured geometric structures may cause undesirable data distribution (\emph{e.g.}, classes are not separable).

In this work, we use the mixed-curvature space to embed data, which is a product of multiple constant curvature spaces acting as submanifolds~\cite{gu2018learning,skopek2019mixed}. 
The mixed-curvature space has shown to be superior to Euclidean space in some machine learning tasks, owing to its ability to capture non-Euclidean geometric structures~\cite{Sun2022ASM,Shevkunov2021OverlappingSF}. 
Examples include image classification~\cite{Gao2022CurvatureAdaptiveMF}, graph analysis~\cite{WangWSWNAXYC21}, and information retrieval~\cite{XuWWLWYDZZXZ22}. 
The geometry of a mixed-curvature space is determined by the number, dimension, and curvature of constant curvature spaces. 
By changing the geometry, we are able to adjust the mixed-curvature space to match specific geometric structures of data~\cite{zhang2021switch}.
For example, positive curvatures are suitable for local cyclical structures, and negative curvatures are suitable for hierarchical structures in a region.
Based on the mixed-curvature space, we restate the two challenges:
\textbf{(1)} how to identify the suitable geometry of the mixed-curvature space for growing geometric structures, and \textbf{(2)} how to preserve the geometric structures of old data when learning from new data in the mixed-curvature space.

We introduce a geometry incremental search scheme to solve the first challenge.
We build a submanifold pool by sampling subsets of coordinates from features, where the length of coordinates is the dimension of constant curvature spaces and features are projected to them using initial curvatures.
Given new data, we select constant curvature spaces that contribute significantly to the current task to expand geometry of the mixed-curvature space. In this case, the growing geometric structures are well encoded.
We introduce two loss functions, \emph{i.e.}, an angular-regularization loss and a neighbor-robustness loss, to solve the second challenge.
The angular-regularization loss penalizes the change of angles between any pair of instances to preserve global structures.
The neighbor-robustness loss realizes within-class compactness and between-class separability in a neighbor to preserve the discriminative power of local structures.
As a result, our method is capable of efficiently learning from new data and preventing forgetting of old data. 
Our method is evaluated on multiple continual learning settings, and
experimental results show the effectiveness of our method.

In summary, our contributions are three-fold.
(1) To the best of our knowledge, we are the first to explore data geometry for continual learning. Our method is efficient for learning from a non-stationary stream of data. 
(2) We introduce an incremental search scheme that identifies the suitable geometry for growing geometric structures of data.
(3) We introduce an angle-regularization loss and a neighbor-robustness loss, capable of preserving geometric structures of old data in the mixed-curvature space.

\section{Related Work}

\subsection{Continual Learning}

Continual learning is mainly studied under two scenarios: task incremental learning~\cite{Kirkpatrick2017OvercomingCF,Zenke2017ContinualLT} and class incremental learning~\cite{Rebuffi2017iCaRLIC,Yu2020SemanticDC}. 
In task incremental learning, a model is trained on incremental tasks with clear boundaries. At the test stage, the task ID is provided. 
In contrast, the class incremental learning targets a more challenging and practical setup, where the task ID is not provided at the test stage. 
In this paper, we focus on class incremental learning.

Existing techniques for continual learning can be broadly divided into three categories: replay-based methods, regularization-based methods, and architecture-based methods. 
Replay-based methods store a small amount of old data~\cite{Rebuffi2017iCaRLIC,Chaudhry2021UsingHT,Buzzega2020DarkEF}, gradients of old data~\cite{LopezPaz2017GradientEM,Chaudhry2019EfficientLL}, features of old data~\cite{Zhu2021PrototypeAA}, or train a generative model for old data~\cite{Shin2017ContinualLW,Zhu2021ClassIncrementalLV}, and review them to tune the model when new data comes. 
Regularization-based methods penalize the update of features or parameters to avoid catastrophic forgetting. Representative schemes include measuring the importance of parameters~\cite{Kirkpatrick2017OvercomingCF,Zenke2017ContinualLT}, projecting gradients onto a null space~\cite{Wang2021TrainingNI,Kong2022BalancingSA}, and distilling knowledge from the old model~\cite{Li2018LearningWF,Tao2020BiObjectiveCL}.
Architecture-based methods allocate specific parameters to different data. When new data comes, some methods select parameters from a fixed super-model~\cite{Mallya2018PackNetAM,Serr2018OvercomingCF} and some methods add new parameters to an expandable model~\cite{Yan2021DERDE,Rusu2016ProgressiveNN}.

Different from existing methods that do not consider data geometry and use Euclidean space for all data, we explore data geometry by producing suitable manifolds for new data and ensuring that the geometric structures are kept unchanged to avoid forgetting.
Compared with preserving features or parameters, preserving geometric structures provides flexibility to the model and makes a better balance between learning new knowledge and remembering old knowledge. Meanwhile, compared with methods that expand the network architecture, expanding geometry of the underlying space in our method avoids huge resource consumption.

\subsection{Mixed-curvature Space}

The mixed-curvature space has shown superior performance to the Euclidean space in various problems such as natural language processing~\cite{nickel2017poincare}, computer vision~\cite{Gao2022CurvatureAdaptiveMF}, and graph analysis~\cite{Bachmann2019ConstantCG}.  
A mixed-curvature space uses multiple constant-curvature spaces as submanifolds to model complex geometric structures.
Despite its success, one challenge that remains is how to match its geometry to the intrinsic structures of data. 
Unsuitable geometry results in overwhelming distortions and inferior performance.
Some methods identify the geometry by trial-and-error experiments~\cite{gu2018learning,skopek2019mixed}, which yet is a resource-intensive process.
Recently, several methods learn to automatically identify the suitable geometry. 
Sun~\emph{et al.}~\cite{Sun2022ASM} utilize the attention mechanism to assign weights for submanifolds.
Wang~\emph{et al.}~\cite{WangWSWNAXYC21} update curvatures of the mixed-curvature space via gradient descent.
Shevkunov and Prokhorenkova~\cite{Shevkunov2021OverlappingSF}, and Zhang~\emph{et al.}~\cite{zhang2021switch} employ overlapping spaces and a sparse gating mechanism to select submanifolds, respectively.

Different from these methods that identify geometry for a fixed geometric structure of data, 
our incremental search scheme is designed for a non-stationary stream of data with growing structures. 
In addition, compared with most methods that use a fixed number of submanifolds and learn curvatures and weights for them, our method dynamically increases the number of constant curvature spaces, capable of matching more complex structures.

\begin{figure*}[!t]
  \centering
  \includegraphics[width=0.92 \textwidth]{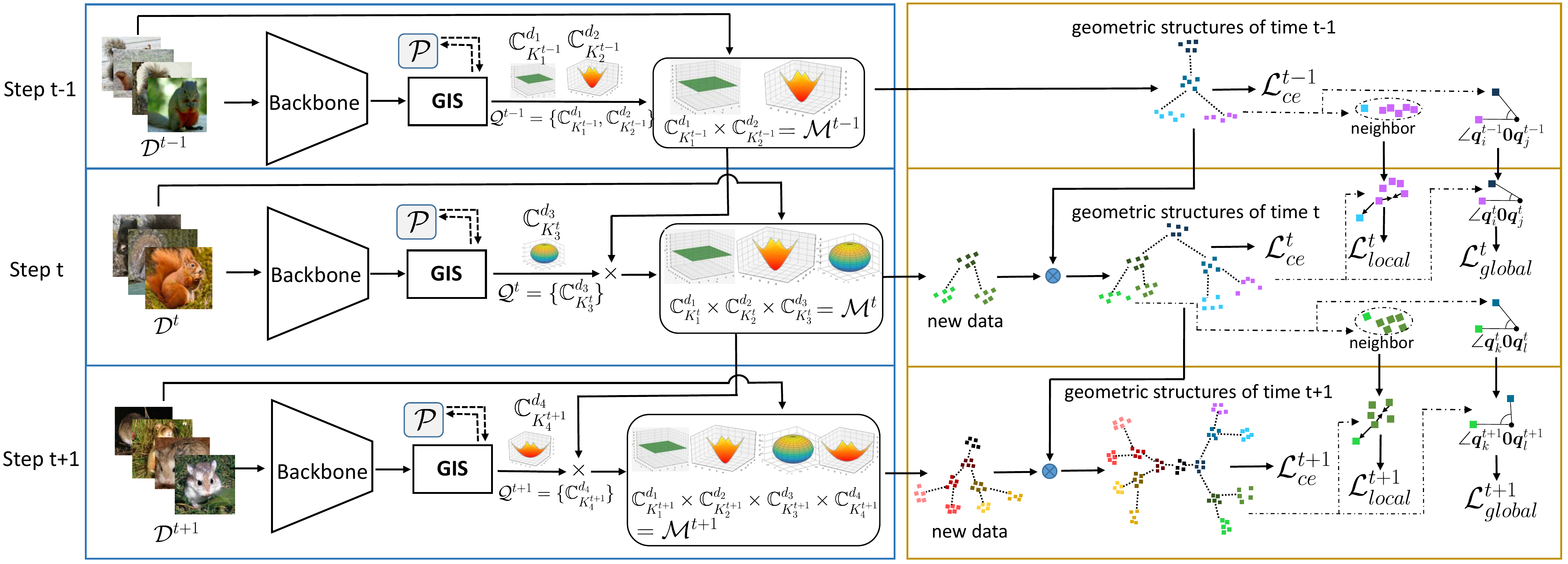}
  \caption{The framework of our method. $\mathcal{D}^t$ denotes the training data given at step $t$. GIS denotes the geometry incremental search scheme, and $\mathcal{P}$ denotes the submanifold pool. $\mathcal{Q}^t$ contains selected CCSs from $\mathcal{P}$ at step $t$, and $\mathcal{M}^t$ is the constructed mixed-curvature space, where $\times$ means the Cartesian product. We train our model under the cross-entropy loss $\mathcal{L}_{ce}^t$, and the newly designed $\mathcal{L}_{global}^t$ and  $\mathcal{L}_{local}^t$ losses.
  }
  \label{fig:framework}
\end{figure*}

\section{Preliminary}

This section describes the constant curvature space and mixed-curvature space involved in our method.

\textbf{Constant curvature space.}
A constant curvature space (CCS) is a smooth Riemannian manifold, denoted by $\mathbb{C}^{d}_{K}$ with the dimension being $d$ and the curvature being $K$. 
The sign of $K$ defines three types of spaces. A negative curvature defines a hyperbolic space, and we use the Poincar{\'e} ball model~\cite{cannon1997hyperbolic} to work for hyperbolic space. A zero curvature defines the Euclidean space. A positive curvature defines a hypersphere space. We opt for the projected sphere model~\cite{buss2001spherical} to work for hypersphere. 
For $\Vec{u} \in \mathbb{C}^{d}_{K}$, its tangent space, denoted by $T_{\Vec{u}} \mathbb{C}^{d}_{K}$, is a Euclidean space. We use the exponential map ${\exp}_{\Vec{u}}^{K}(\cdot): T_{\Vec{u}} \mathbb{C}^{d}_{K} \to \mathbb{C}^{d}_{K}$ and the logarithmic map ${\log}_{\Vec{u}}^{K}(\cdot): \mathbb{C}^{d}_{K} \to T_{\Vec{u}} \mathbb{C}^{d}_{K}$ to achieve transformations between CCS and its tangent space,
\begin{equation}
\resizebox{.95\columnwidth}{!}{$
\label{equation:exponentialmap}
\left\{
\begin{aligned}
& {\exp}_{\Vec{u}}^{K}(\Vec{q}) = \Vec{u} {\oplus}_{K} \Big( {\tan}_{K} (\sqrt{| K |} \frac{\lambda_{\Vec{u}}^{K} \| \Vec{q} \|  }{2}   )   \frac{\Vec{q}}{\sqrt{|K|}  \cdot  \| \Vec{q} \| } \Big) \\
& {\log}_{\Vec{u}}^{K}(\Vec{x}) = \frac{2}{\sqrt{| K|} \lambda_{\Vec{u}}^{K}  }   {\tan}_{K}^{-1} (\sqrt{|K|} \cdot \| -\Vec{u} {\oplus}_{K} \Vec{x}  \|  )  \frac{ -\Vec{u} {\oplus}_{K} \Vec{x} }{ \| -\Vec{u} {\oplus}_{K} \Vec{x}  \|} 
\end{aligned}
\right.,
$}
\end{equation}
where, ${\oplus}_{K}$ is the M{\"o}bius addition operation, defined as
\[ \small \Vec{x} {\oplus}_{K} \Vec{y} = \frac{ (1-2K \langle \Vec{x}, \Vec{y}  \rangle_{2} - K \|  \Vec{y} \|^{2}) \Vec{x} + (1 + K\| \Vec{x} \|^{2}  ) \Vec{y}  }   {1-2K \langle \Vec{x}, \Vec{y} \rangle_{2} + K^{2} \| \Vec{x} \|^{2} \| \Vec{y} \|^{2}}\;, \]
and ${\tan}_{K}(\cdot)$ is $\tan(\cdot)$ if $K \geq 0$, otherwise ${\tan}_{K}(\cdot)$ is the hyperbolic tangent function $\tanh(\cdot)$.  
Similarly, ${\tan}_{K}^{-1}(\cdot) = {\tan}^{-1}(\cdot)$ if $K \geq 0$, otherwise ${\tan}_{K}^{-1}(\cdot) = {\tanh}^{-1}(\cdot)$. 

A CCS is equipped with a metric $g: {T}_{\Vec{u}} \mathbb{C}^{d}_{K} \times {T}_{\Vec{u}} \mathbb{C}^{d}_{K} \to \mathbb{R}$ that induces the distance and angle on the space.
For $ \Vec{x}, \Vec{y} \in \mathbb{C}^{d}_{K}$, distance between them is
\begin{equation}
\label{equation:distancemeasure}
\small
\begin{aligned}
{\psi}_{K}(\Vec{x},\Vec{y}) = \frac{2}{\sqrt{|K|}} {\tan}_{K}^{-1} ( \sqrt{ | K | }  \cdot  \| -\Vec{x} {\oplus}_{K} \Vec{y}  \|   ).
\end{aligned}
\end{equation}
The angle in the CCS at the origin $\Mat{0}$ is conformal to that in Euclidean space. For vectors $ \Vec{q}, \Vec{s} \in T_{\Vec{0}} \mathbb{C}^{d}_{K}$, the angle between them is
\begin{equation}
\small
\label{equation:angle}
\begin{aligned}
\cos (\angle\Vec{q}\Mat{0}\Vec{s} ) = \frac{g(\Vec{q},\Vec{s})}{\sqrt{ g(\Vec{q},\Vec{q})} \sqrt{ g(\Vec{s},\Vec{s})}  }  = \frac{\langle \Vec{q},\Vec{s} \rangle}{\| \Vec{q}\| \| \Vec{s}\|}. 
\end{aligned}
\end{equation}

\textbf{Mixed-curvature space.}
The mixed-curvature space is defined as the Cartesian product of multiple CCSs, $\mathcal{M} \coloneqq  {\times}_{j=1}^{m} \mathbb{C}^{d_j}_{K_j} $~\cite{gu2018learning}, where $\mathbb{C}^{d_j}_{K_j}$ is the $j$-th CCS, acting as the $j$-th submanifold. There are $m$ CCSs in $\mathcal{M}$ totally. 
Curvatures of all CCSs are denoted by $\mathcal{K} = \{ K_1,\cdots,K_m \}$. 
Any $\Vec{x} \in \mathcal{M}$ is represented by a vector concatenation, $\Vec{x} = (\Vec{x}_{1},\cdots,\Vec{x}_{m})$, where $\Vec{x}_{j} \in \mathbb{C}^{d_j}_{K_j}$. 
Similarly, the tangent space $T_{\Vec{u}}  \mathcal{M}$ of $\mathcal{M}$ is defined by the Cartesian product of tangent spaces of CCSs, $T_{\Vec{u}}  \mathcal{M} \coloneqq  {\times}_{j=1}^{m} T_{\Vec{u}_j}  \mathbb{C}^{d_j}_{K_j}$, with $\Vec{u}= (\Vec{u}_1,\cdots,\Vec{u}_m) \in \mathcal{M}$ being the tangent point. A tangent vector $\Vec{q} \in  T_{\Vec{u}}  \mathcal{M}$ is $\Vec{q} = (\Vec{q}_{1},\cdots,\Vec{q}_{m})$, where $\Vec{q}_{j} \in T_{\Vec{u}_j}  \mathbb{C}^{d_j}_{K_j}$. 
The exponential map ${\bexp}_{\Vec{u}}^{\mathcal{K}}(\cdot): T_{\Vec{u}} \mathcal{M} \to \mathcal{M}$ and the logarithmic map ${\blog}_{\Vec{u}}^{\mathcal{K}}(\cdot): \mathcal{M} \to T_{\Vec{u}} \mathcal{M}$ in the mixed-curvature space is 
\begin{equation}
\small
\label{equation:log}
\left\{
\begin{aligned}
& \bexp_{\Vec{u}}^{\mathcal{K}}(\Vec{q}) = \Big( \exp_{\Vec{u}_1}^{K_1}(\Vec{q}_1),\cdots,\exp_{\Vec{u}_m}^{K_m}(\Vec{q}_m) \Big) \\
& \blog_{\Vec{u}}^{\mathcal{K}}(\Vec{x}) = \Big( \log_{\Vec{u}_1}^{K_1}(\Vec{x}_1),\cdots,\log_{\Vec{u}_m}^{K_m}(\Vec{x}_m) \Big) 
\end{aligned}
\right..
\end{equation}

If the $j$-th CCS has the metric $g_{j}$, the mixed-curvature space has the metric $G(\Vec{q},\Vec{s}) = \sum_{j=1}^{m} g_{j} (\Vec{q}_j,\Vec{s}_j)$. 
For two vectors $\Vec{x}, \Vec{y} \in \mathcal{M}$, the squared distance between them in the mixed-curvature space is given by the sum of squared distances in CCSs,
\begin{equation}
\label{equation:distancemeasuremixed}
\begin{aligned}
{\Psi}^2(\Vec{x},\Vec{y}) = {\sum}_{j=1}^{m} {\psi}_{K_{j}}^2 (\Vec{x}_{j},\Vec{y}_{j}),
\end{aligned}
\end{equation}
where $\psi_{K_{j}}^2 (\cdot)$ is the squared distance in $\mathbb{C}^{d_j}_{K_j}$.
For two vectors $\Vec{q}, \Vec{s} \in  T_{\Vec{0}}  \mathcal{M}$, the angle between them is 
\begin{equation}
\label{equation:anglemixed}
\resizebox{9 cm}{!}{$
\begin{aligned}
& \cos (\angle\Vec{q}\Vec{0}\Vec{s} ) = \frac{G(\Vec{q},\Vec{s})}{\sqrt{ G(\Vec{q},\Vec{q})} \sqrt{ G(\Vec{s},\Vec{s})}  }  = \frac{ \sum_{j=1}^{m} g_{j} (\Vec{q}_j,\Vec{s}_j)}{ \sqrt{\sum_{j=1}^{m} g_{j} (\Vec{q}_j,\Vec{q}_j)}  \sqrt{\sum_{j=1}^{m} g_{j} (\Vec{s}_j,\Vec{s}_j)}  } = \frac{\langle \Vec{q},\Vec{s} \rangle}{\| \Vec{q}\| \| \Vec{s}\|}.
\end{aligned}
$}
\end{equation}

\section{Method}

\subsection{Formulation}

Continual learning is cast as training a model from a stream of data $(\mathcal{D}^{1}, \cdots, \mathcal{D}^{T})$, received at $T$ distinct steps. 
$\mathcal{D}^{t}$ denotes the training data at step $t$, containing $N_t$ pairs of the
form of $(\Mat{I},{y})$,
where $\Mat{I}$ is an instance (\emph{e.g.}, an image) and ${y}$ is the  associated label from the label space $\mathcal{Y}^t$. Label spaces at different steps have no overlap, $\mathcal{Y}^1 \cap \cdots \cap \mathcal{Y}^T=\emptyset$.
$\mathcal{D}^{t}$ is only available at step $t$. Our goal is to correctly classify instances from all seen classes.
In other words, after the training process at step $t$, the model is evaluated in the label space $\cup_{i=1}^t \mathcal{Y}^i$.

We use a backbone $f_{\theta^t}(\cdot)$ to extract features from an instance $\Mat{I}$, where $\theta^t$ denotes the parameter at step $t$.
Then, $f_{\theta^t}(\Mat{I}) \in \mathbb{R}^d $ is projected to $\Vec{x}^{t}=(\Vec{x}_1^{t},\cdots,\Vec{x}_{m_t}^{t})$ in the mixed-curvature space $\mathcal{M}^{t}$, where $\mathcal{M}^{t}={\times}_{j=1}^{m_t} \mathbb{C}^{d_j}_{K_j^t}$ is produced by our geometry incremental search scheme ${\rm GIS}(\mathcal{M}^{t-1}, \mathcal{D}^t)$ that expands $\mathcal{M}^{t-1}$ to get $\mathcal{M}^{t}$ using $\mathcal{D}^t$, and $m_t$ is the number of CCSs.

We perform distance-based classification for $\Vec{x}^{t}$.
At step $t$, suppose that there are $n$ classes totally and the classifier of the $l$-th class is $\Vec{W}_l^t \in \mathbb{R}^d$, classifiers of all classes are denoted by $\Vec{W}^t =[\Vec{W}_1^t,\cdots,\Vec{W}_n^t]$.
$\Vec{W}_l^t$ is also projected to $\Vec{w}_l^t \in \mathcal{M}^{t}$ in the mixed-curvature space.
The probability that $\Vec{x}^{t}$ belongs to the $l$-th class is
\begin{equation}
\label{equation:ourclassification}
\small
\begin{aligned}
& p(l| {\Vec{x}^{t}}) =   \frac{  \exp \big( -{\Psi}^2 ( \Vec{x}^{t},\Vec{w}_l^t) \big)   }   {  \sum_{l'=1}^{n}  \exp \big(-{\Psi}^2 (   \Vec{x}^{t}, \Vec{w}_{l'}^t  ) \big) } .
\end{aligned}
\end{equation}

We update the backbone and the classifier by minimizing the cross-entropy loss $\mathcal{L}_{ce}^t$ for classification, and the newly designed angular-regularization loss $\mathcal{L}_{global}^t$ and neighbor-robustness loss $\mathcal{L}_{local}^t$ for preventing the change of geometric structures.
Our framework is shown in~\cref{fig:framework}.

\begin{figure*}[!t]
  \centering
  \includegraphics[width= 0.9 \textwidth]{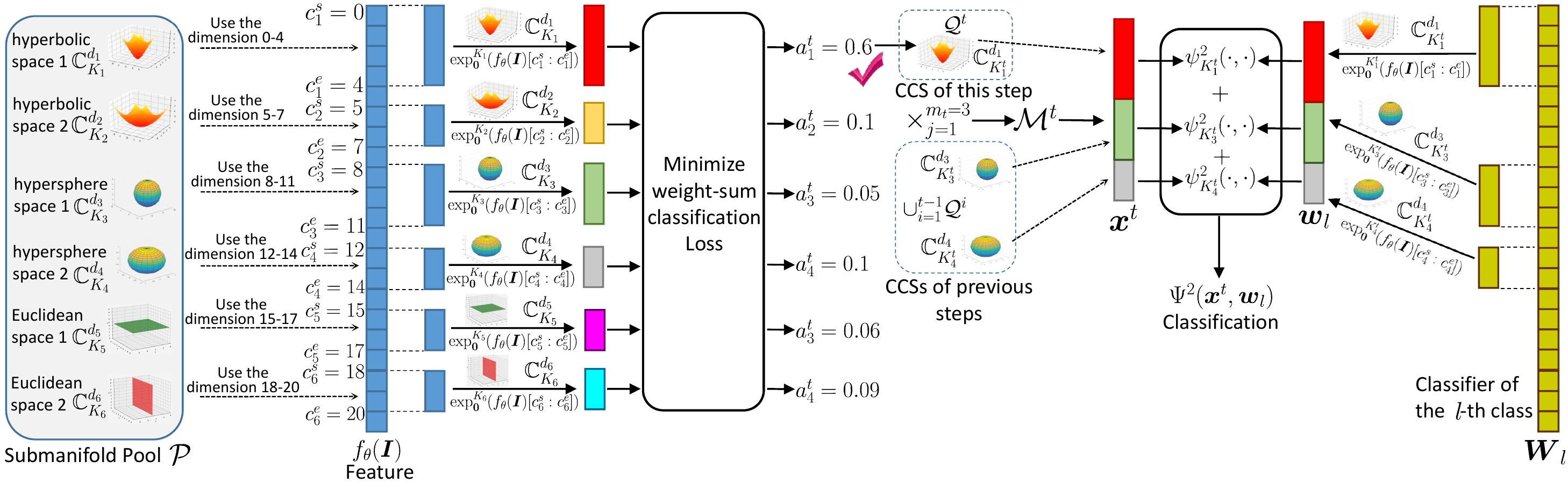}
  \caption{The illustration of the geometry incremental search. We build a submanifold pool $\mathcal{P}$ of six CCSs. We project feature $f_{\theta}(\Mat{I}) \in \mathbb{R}^d$ to CCSs, and update weights $a_j^t$ and curvatures $K_j^t$ for CCSs by minimizing a weight-sum classification loss.
  The mixed-curvature space $\mathcal{M}^{t}$ is constructed by combining CCSs with high weights (contained in $\mathcal{Q}^t$) and CCSs selected at previous steps (contained in $\cup_{i=1}^{t-1}\mathcal{Q}^i$).
  We also project the classifier $\Mat{W}_l \in \mathbb{R}^d$ to $\Mat{w}_l \in \mathcal{M}^{t}$. 
  The distance $\Psi^2(\Mat{x}_t,\Mat{w}_l)$ composed of distances $\psi^2_{K_j^t}$ in CCSs is used for classification.
  }
  \label{fig:geometry}
\end{figure*}

\subsection{Geometry Incremental Search Scheme}

\noindent \textbf{Notation.} 
At step $t$, we omit the index $t$ of the backbone $f_{\theta}$, classifier $\Mat{W}$, and curvature $K$ if they are not trained and initialized by results from step $t-1$.
For example, at step $t$, we use $\theta$ to denote the backbone parameter that is initialized by $\theta^{t-1}$, and $\theta^{t}$ means it is well-trained at step $t$.

Before training, we build a submanifold pool of CCSs, $\mathcal{P} = \{\mathbb{C}^{d_1}_{K_1},\mathbb{C}^{d_2}_{K_2},\cdots,\mathbb{C}^{d_{\xi}}_{K_{\xi}} \}$, by sampling subsets of coordinates from $f_{\theta}(\Mat{I}) \in \mathbb{R}^{d}$. 
Concretely,

\noindent (1) Each $\mathbb{C}^{d_j}_{K_j}$ corresponds to a subset of coordinates in $f_{\theta}(\Mat{I})$, starting from the $c_j^s$-th dimension to the $c_j^e$-th dimension in $f_{\theta}(\Mat{I})$, and the dimension $d_j = c_j^e - c_j^s + 1$. The correspondence rule (\emph{i.e.}, $c_j^s$ and $c_j^e$) is pre-defined and does not change with $t$. For example, we can randomly generate them with $c_j^s<c_j^e<d$.

\noindent (2) Representation on $\mathbb{C}^{d_j}_{K_j}$ is $\Vec{x}_j = {\exp}_{\Vec{0}}^{K_j}(f_{\theta}(\Mat{I})[c_j^s:c_j^e])$, where $f_{\theta}(\Mat{I})[c_j^s:c_j^e]$ is a $d_j$-dimension vector.  

\noindent (3) The number $\xi$ of CCSs in $\mathcal{P}$ does not change with $t$.

\noindent (4) Curvatures $\mathcal{K} = \{K_1,\cdots,K_{\xi}\}$ change with $t$. At the beginning of step $t$, $\mathcal{K}$ is initialized by curvatures from step $t-1$. (Curvatures are then updated via Eq.~\eqref{equation:lossfunctionweight})

At step $t$, we select CCSs from $\mathcal{P}$ to expand the geometry of the underlying space, according to their importance for classifying $\mathcal{D}_{t}$.
We project the classifier $\Vec{W}_l$ of the $l$-th class to these CCSs based on sampled coordinates, $\Vec{w}_{lj} ={\exp}_{\Vec{0}}^{K_j}(\Mat{W}_l[c_j^s:c_j^e])$.
We assign initial weight $a_j$ to the $j$-th CCS, and all weights are collectively denoted by $\mathcal{A}=\{ a_1, \cdots, a_m\}$. 
To identify proper CCSs, we minimize the following weight-sum classification loss,
\begin{equation}
\label{equation:lossfunctionweight}
\resizebox{8.5 cm}{!}{$
\begin{aligned}
\mathcal{A}^{t},\mathcal{K}^{t} = \arg\min_{\mathcal{A},\mathcal{K}}  {\mathbb{E}}_{\Vec{I} \sim \mathcal{D}} \left[ - \log \frac{  \exp \big( -\sum_{j=1}^{m} a_j \cdot {\psi}_{K_j}^2 (\Vec{x}_j, \Vec{w}_{{{y}}j}) \big)   }   {  \sum_{l'=1}^{n}  \exp \big(-\sum_{j=1}^{m} a_j \cdot {\psi}_{K_j}^2 (\Vec{x}_j, \Vec{w}_{l'j})  \big) } \right], 
\end{aligned}
$}
\end{equation}
where ${{y}}$ is the ground truth of $\Vec{I}$, $\mathcal{A}^{t}= \{a^{t}_1,\cdots,a^{t}_m\}$, and $\mathcal{K}^{t}= \{K^{t}_1,\cdots,K^{t}_m\}$.
We select CCSs from $\mathcal{P}$ according to their weights $a_j^t$, 
\begin{equation}
\label{equation:submanifoldselect}
\small
\begin{aligned}
\mathcal{Q}^t = \{\mathbb{C}^{d_j}_{K^{t}_j} | a^{t}_j > \tau_1 \},
\end{aligned}
\end{equation}
where $\tau_1$ is a threshold. 
The geometry incremental search scheme expands $\mathcal{M}^{t-1}$ to get $\mathcal{M}^{t}$ by 
combining $\mathcal{Q}^t$ with CCSs selected at previous steps,
\begin{equation}
\label{equation:submanifoldselect2}
\small
\begin{aligned}
\mathcal{M}^{t} := {\times}_{\mathbb{C}^{d}_{K} \in {\cup}_{i=1}^{t} \mathcal{Q}^i} \mathbb{C}^{d}_{K},
\end{aligned}
\end{equation}
where $| {\cup}_{i=1}^{t} \mathcal{Q}^i| = m_t$. 
The representation in $\mathcal{M}^{t}$ is $\Vec{x}^{t} =(\Vec{x}_1^{t},\cdots,\Vec{x}_{m_t}^{t})$, where $\Vec{x}_j^{t} =  {\exp}_{\Vec{0}}^{K^{t}_j}(f_{\theta}(\Mat{I})[c_j^s:c_j^e]) \in \mathbb{C}^{d_j}_{K^{t}_j}$, and $\mathbb{C}^{d_j}_{K^{t}_j} \in {\cup}_{i=1}^{t} \mathcal{Q}^i$. 
Similarly, the classifier for the $l$-th task is $\Vec{w}_{l} =(\Vec{w}_{l1},\cdots,\Vec{w}_{lm_t}) \in \mathcal{M}^{t}$, where $\Vec{w}_{lj} =  {\exp}_{\Vec{0}}^{K^{t}_j}(\Mat{W}_l[c_j^s:c_j^e]) \in \mathbb{C}^{d_j}_{K^{t}_j}$, and $\mathbb{C}^{d_j}_{K^{t}_j} \in {\cup}_{i=1}^{t} \mathcal{Q}^i$. 
Based on $\mathcal{M}^{t}$, we train $\theta$ and $\Mat{W}$ to maximize the performance.
Note that, although the dimension of $\Vec{x}^{t}$ may grow with $t$, the dimension of both the feature $f_{\theta}(\Mat{I})$ and the classifier $\Mat{W}_l$ are fixed as $d$. 
The architecture and parameter number of the backbone are fixed during training, and we just add classifiers for new classes. Thus, our resource consumption does not grow much in the data stream. 
This process is shown in~\cref{fig:geometry}.

\subsection{Geometric Structure Preserving}

To alleviate catastrophic forgetting, we preserve captured geometric structures of old data, through which instances of the same class are close to each other and far from those of different classes.
In doing so, we store a few instances of old classes in a memory buffer $\mathcal{B}$, and introduce the angular-regularization loss $\mathcal{L}_{global}$ for global structures and the neighbor-robustness loss $\mathcal{L}_{local}$ for local structures.

\subsubsection{Angular regularization loss}
We use angles between instances to characterize the global geometric structures of data.
The idea is that if angles between old data are preserved, the geometric structures are also preserved.
At step $t$, we measure angles of instances in the memory buffer, and changes of the angles compared with those at step $t-1$ are penalized.

Concretely, for any two instances $\Mat{I}, \Mat{I}' \in \mathcal{B}$, we obtain their representations $\Vec{x}^{t}, \Vec{x}'^{t} \in \mathcal{M}^{t}$. 
Their representations $\Vec{x}^{t-1}, \Vec{x}'^{t-1} \in \mathcal{M}^{t-1}$ are also computed using $f_{\theta^{t-1}}$ and $\mathcal{M}^{t-1}$.
We project $\Vec{x}^{t}, \Vec{x}'^{t}, \Vec{x}^{t-1}, \Vec{x}'^{t-1}$ to tangent spaces,
\begin{equation}
\label{equation:tangentspace}
\small
\left\{ 
\begin{aligned}
& \Vec{q}^{t} = \blog_{\Mat{0}}^{\mathcal{K}^{t}}(\Vec{x}^{t}) \\
& \Vec{q}'^{t} = \blog_{\Mat{0}}^{\mathcal{K}^{t}}(\Vec{x}'^{t}) \\
\end{aligned}
\right.,
\left\{ 
\begin{aligned}
& \Vec{q}^{t-1} = \blog_{\Mat{0}}^{\mathcal{K}^{t-1}}(\Vec{x}^{t-1})\\
& \Vec{q}'^{t-1} = \blog_{\Mat{0}}^{\mathcal{K}^{t-1}}(\Vec{x}'^{t-1}) 
\end{aligned}
\right.,
\end{equation}
where $\Vec{q}^{t},\Vec{q}'^{t} \in  T_{\Mat{0}} \mathcal{M}^{t}$ and $\Vec{q}^{t-1},\Vec{q}'^{t-1} \in  T_{\Mat{0}} \mathcal{M}^{t-1}$.
Their angles in the tangent spaces are computed by
\begin{equation}
\small
\label{equation:oldangle}
\begin{aligned}
& \cos (\angle \Vec{q}^{t}  \Vec{0}  \Vec{q}'^{t}) = \frac{\langle \Vec{q}^{t}, \Vec{q}'^{t} \rangle }{\|\Vec{q}^{t}\|  \| \Vec{q}'^{t}\|}, \\
& \cos (\angle \Vec{q}^{t-1} \Vec{0} \Vec{q}'^{t-1}) = \frac{ \langle \Vec{q}^{t-1},\Vec{q}'^{t-1} \rangle }{\|\Vec{q}^{t-1}\|  \|  \Vec{q}'^{t-1}\|}, \\
\end{aligned}
\end{equation}
The angular-regularization loss penalizes the difference between $\angle \Vec{q}^{t}  \Vec{0}  \Vec{q}'^{t}$ and $\angle \Vec{q}^{t-1} \Vec{0} \Vec{q}'^{t-1}$ by 
\begin{equation}
\small
\label{equation:bloballoss}
\begin{aligned}
\mathcal{L}_{global}^{t} = \sum_{(\Mat{I}_i, \Mat{I}_j \in \mathcal{B})}  l_{\delta} \big( \cos (\angle \Vec{q}^{t} \Vec{0} \Vec{q}'^{t}), \cos (\angle \Vec{q}^{t-1} \Vec{0} \Vec{q}'^{t-1}) \big),
\end{aligned}
\end{equation}
where $l_{\delta}$ is the Huber loss~\cite{ParkKLC19}, 
\begin{equation}
\small
\label{equation:huberloss}
\begin{aligned}
l_{\delta} (a,b) = \left\{ 
\begin{aligned}
& \frac{1}{2} (a-b)^2, \  {\rm if} \ |a-b|\leq 1,\\
& |a-b| - \frac{1}{2}, \  {\rm otherwise}
\end{aligned}
\right..
\end{aligned}
\end{equation}

\subsubsection{Neighbor-robustness loss}

The neighbor-robustness loss preserves the local structures by realizing within-class compactness and between-class separability. 
At step $t$, for any $\Mat{I}, \Mat{I}' \in \mathcal{B}$, the squared distance ${\Psi}^2(\Vec{x}^{t-1}, \Vec{x}'^{t-1})$ between their representations $\Vec{x}^{t-1}$ and $\Vec{x}'^{t-1}$ are computed.

We measure neighbors of $\Mat{I}$ at step $t-1$ as
\begin{equation}
\small
\label{equation:neighbor}
\left\{ 
\begin{aligned}
& \mathcal{N}^{t-1}_w(\Mat{I}) =\{\Mat{I}' | {\Psi}^2(\Vec{x}^{t-1}, \Vec{x}'^{t-1}) < \tau_2 \ {\rm and} \ {y}={y}' \} \\
& \mathcal{N}^{t-1}_b(\Mat{I}) =\{\Mat{I}' | {\Psi}^2(\Vec{x}^{t-1}, \Vec{x}'^{t-1}) < \tau_2 \ {\rm and} \ {y} \neq {y}' \} 
\end{aligned}
\right.,
\end{equation}
where $\mathcal{N}^{t-1}_w(\Mat{I})$ is the within-class neighbors of $\Mat{I}$, $\mathcal{N}^{t-1}_b(\Mat{I})$ denotes the between-class neighbors of $\Mat{I}$, and $\tau_2$ is a threshold.
Next, we calculate the within-class and between-class coefficients between $\Mat{I}, \Mat{I}'$ as
\begin{equation}
\small
\label{equation:withinclassneighbors}
w^{t-1} = \begin{cases}
1, &\mbox{ if } \ \Mat{I} \in \mathcal{N}^{t-1}_w(\Mat{I}') \ \mbox{or} \ \Mat{I}' \in \mathcal{N}^{t-1}_w(\Mat{I}) \\
0, &\mbox{otherwise}
\end{cases}
,
\end{equation}
\begin{equation}
\small
\label{equation:betweenclassneighbors}
b^{t-1} = \begin{cases}
1, &\mbox{ if } \ \Mat{I} \in \mathcal{N}^{t-1}_b(\Mat{I}') \ \mbox{or} \ \Mat{I}' \in \mathcal{N}^{t-1}_b(\Mat{I}) \\
0, &\mbox{otherwise}
\end{cases}
\;.
\end{equation}
Based on $w^{t-1}$ and $b^{t-1}$, an affinity coefficient is computed by $e^{t-1} = w^{t-1} - b^{t-1}$. We summarize the above process as $e^{t-1} = E\big(\mathcal{N}^{t-1}_w(\Mat{I}), \mathcal{N}^{t-1}_w(\Mat{I}') \big)$.

Finally, the neighbor-robustness loss forces the squared distance $\Psi^2( \Vec{x}^{t}, \Vec{x}'^{t})$ at step $t$ to comply with the affinity coefficient at step $t-1$,
\begin{equation}
\label{equation:localloss}
\begin{aligned}
\mathcal{L}_{local}^{t} & = \sum_{(\Mat{I}, \Mat{I}' \in \mathcal{B})}   E\big( \mathcal{N}^{t-1}_w(\Mat{I}), \mathcal{N}^{t-1}_w(\Mat{I}')\big) {\Psi}^2( \Vec{x}^{t}, \Vec{x}'^{t}),
\end{aligned}
\end{equation}
through which local structures become discriminative.

\begin{algorithm}
 \small
 \caption{Training process of the proposed method.}
 \label{algorithm:Updateparameters}
 \begin{algorithmic}[1]
 \renewcommand{\algorithmicrequire}{\textbf{Input:}}
 \renewcommand{\algorithmicensure}{\textbf{Output:}}
 \REQUIRE Data stream $(\mathcal{D}^{1}, \cdots, \mathcal{D}^{T})$. Randomly initialized backbone $f_{\theta}$ and classifier $\Vec{W}$. Memory buffer $\mathcal{B} = \emptyset$.
 \ENSURE  Updated backbone $f_{\theta^{T}}$ and classifier $\Vec{W}^{T}$.
  \WHILE { $t < T$}
    \STATE Initialize ${\theta}$ and $\Vec{W}$ by parameters from step $t-1$.
    \WHILE {$k < $ MaxIteration }
      \STATE Optimize weights $\mathcal{A}$ and curvatures $\mathcal{K}$ of CCSs via Eq.~\eqref{equation:lossfunctionweight}, using data from $\mathcal{D}^{t}$.
    \ENDWHILE
    \STATE Produce the underlying space $\mathcal{M}^{t}$ for $\mathcal{D}^{t}$ via the geometry incremental search scheme via Eq.~\eqref{equation:submanifoldselect} and Eq.~\eqref{equation:submanifoldselect2}. 
    \WHILE {$k < $ MaxIteration }
      \STATE Update the backbone $f_{\theta}$ and classifier $\Vec{W}$ by minimizing the classification loss $\mathcal{L}_{ce}^t$, angular-regularization loss $\mathcal{L}_{global}^t$, and neighbor-discriminative loss $\mathcal{L}_{local}^t$ in Eq.~\eqref{equation:totalloss}.
    \ENDWHILE
    \STATE Obtain $f_{\theta^{t}}$ and $\Vec{W}^{t}$.
    \STATE Randomly select a few instances from $\mathcal{D}^{t}$ and add them to $\mathcal{B}$.
  \ENDWHILE
  \STATE Return $f_{\theta^{T}}$ and $\Vec{W}^{T}$.
 \end{algorithmic}
\label{algorithm:training}
\end{algorithm}

\subsection{Training}

Our goal is to train the parameter $\theta$ of the backbone $f_{\theta}$ and the classifier $\Vec{W}$ in the stream of data.
Given new data $\mathcal{D}^t$ at step $t$, we first produce the mixed-curvature space $\mathcal{M}^t$ via the geometry incremental search scheme.
Then, we train $\theta$ and $\Vec{W}$ under the guidance of a cross-entropy loss $\mathcal{L}_{ce}^t$, the angular-regularization loss $\mathcal{L}_{global}^t$, and the neighbor-robustness loss $\mathcal{L}_{local}^t$.
The cross-entropy loss $\mathcal{L}_{ce}^t$ is 
\begin{equation}
\label{equation:corssentropy}
\begin{aligned}
\mathcal{L}_{ce}^t = {\mathbb{E}}_{\Mat{I} \sim (\mathcal{D}^t \cup \mathcal{B}) }  \left[ - \log p({\widehat{y}}| \Mat{I}) \right].
\end{aligned}
\end{equation}
Overall, the loss function at step $t$ is given by
\begin{equation}
\label{equation:totalloss}
\begin{aligned}
\mathcal{L}^t = \mathcal{L}_{ce}^t + \lambda_1  \mathcal{L}_{global}^t + \lambda_2  \mathcal{L}_{local}^t,
\end{aligned}
\end{equation}
where $\lambda_1$ and $\lambda_2$ are the trade-off hyperparameters. 
After the training process, we obtain the backbone $\theta^t$ and the classifier $\Mat{W}^t=[\Vec{W}_{1}^{t},\cdots,\Vec{W}_{n}^{t}]$. Then we randomly select a few instances from $\mathcal{D}^t$ and add them into $\mathcal{B}$.
The pseudo-code of the training process is summarized in Algorithm~\ref{algorithm:training}.

\section{Experiments}

\subsection{Settings}

\textbf{Datasets.} 
We evaluate our method on the CIFAR-100~\cite{Krizhevsky2009LearningML} and Tiny-ImageNet~\cite{Le2015TinyIV} datasets that are commonly used in continual learning. 
CIFAR-100 has $100$ classes, and Tiny-ImageNet consists of $200$ classes.

We test our method on two groups of settings following standard protocols: whether the model is pre-trained on a large number of classes~\cite{Zhu2021PrototypeAA,Wang2022ContinualLW}.
(1) We pre-train our model in a large number of classes. For CIFAR-100, we test our method on three settings: C100-B50-S5, C100-B50-S10, and C100-B40-S20. For Tiny-ImageNet, we test our method  on three settings: T200-B100-S5, T200-B100-S10, and T200-B100-S20.
For instance, C100-B50-S5 means that we first pre-train the model using the first $50$ classes, and the following classes are split into $5$ steps that each has $10$ classes. 
(2) We do not pre-train our model. We use the CIFAR-100 dataset and test our method on three settings: C100-B0-S5, C100-B0-S10, C100-B0-S20. 
The second group of settings are more challenging.

\textbf{Experimental Details.}
ResNet-18~\cite{He2016DeepRL} is used as the backbone and trained from scratch in our experiments.
In the training process, we set the batchsize as $64$ and use the Adam optimizer with $0.001$ initial learning rate.
We train the model for $30$ epochs for each step. 
We set the threshold $\tau_1$ as $\frac{1}{n}$ for the geometry incremental search scheme, where $n$ is the number of classes.
For the threshold $\tau_2$, we set it as the mean of squared distances of instances with the same class.
The size of the memory buffer is the same as existing replay-based  methods~\cite{Rebuffi2017iCaRLIC,CastroMGSA18,HouPLWL19}.
For the first group of settings, we randomly select $20$ instances for each class.
For the second group of settings, we evaluate our method with the size of the memory buffer $\mathcal{B}$ as $200$ and $500$.

In the submanifold pool $\mathcal{P}$, CCSs are sampled with size $16$, $32$, $64$, $128$, and $256$.
For simplicity, we sequentially sample coordinates from features for CCSs with the same size. 
Take CCSs with a dimension of $16$ as an example, the first CCS uses dimensions $1-16$, and the second one uses dimensions $17-32$. 
Since the dimension of features from ResNet-18 is $512$, there are $\frac{512}{16} + \frac{512}{32} +\frac{512}{64}+\frac{512}{128}+\frac{512}{256} = 62 $ CCSs in $\mathcal{P}$ totally. 
We initialize curvatures of half of CCSs as $-1$, and initialize curvatures of the rest of CCSs as $1$.
The initial weight $a_j$ is assigned as $\frac{1}{n}$.

\textbf{Evaluation Metric.}
We use the final accuracy, average accuracy, average forgetting, and average incremental accuracy~\cite{Zhu2021ClassIncrementalLV,CastroMGSA18} to evaluate our method.

\begin{table}
\centering
\resizebox{1\columnwidth}{!}{
\begin{tabular}{c|c|c|c}
\hline
 Method &  C100-B50-S5 &  C100-B50-S10 &  C100-B40-S20  \\
\hline
iCaRLCNN~\cite{Rebuffi2017iCaRLIC} & $40.19$ & $38.87$ & $34.26$\\
iCaRLNME~\cite{Rebuffi2017iCaRLIC} & $49.14$ & $45.31$ & $40.53$\\
EEIL~\cite{CastroMGSA18} & $50.21$ & $47.60$ & $42.23$\\
LUCIR~\cite{HouPLWL19} & $54.71$ & $50.53$ & $48.00$\\
PASS~\cite{Zhu2021PrototypeAA} & $55.67$ & $49.03$ & $48.48$\\
IL2A~\cite{Zhu2021ClassIncrementalLV} & $54.98$ & $45.07$ & $45.74$\\
HFA~\cite{HFAGAOZHI2022} & $55.55$ & $52.11$& $46.12$\\
\hline
\bfseries Ours & $\mathbf{56.03}$ & $\mathbf{54.31}$ & $\mathbf{49.32}$ \\
\hline
\end{tabular}
}
\caption{Final accuracy ($\%$) on the C100-B50-S5, C100-B50-S10, and C100-B40-S20 settings.}
\label{table:cifarcontinuallearning}
\end{table}

\begin{table}
\centering
\resizebox{1\columnwidth}{!}{
\begin{tabular}{c|c|c|c}
\hline
 Method & T200-B100-S5 & T200-B100-S10 &  T200-B100-S20  \\
\hline
iCaRLCNN~\cite{Rebuffi2017iCaRLIC} & $23.17$ & $20.71$ & $20.28$\\
iCaRLNME~\cite{Rebuffi2017iCaRLIC} & $34.43$ & $33.24$ & $27.51$\\
EEIL~\cite{CastroMGSA18} & $35.00$ & $33.67$ & $27.64$\\
IL2A~\cite{Zhu2021ClassIncrementalLV} & $36.58$ & $34.28$ & $28.34$\\
HFA~\cite{HFAGAOZHI2022} & $36.11$ & $33.65$ & $31.37$ \\
\hline
\bfseries Ours & $\mathbf{38.10}$ & $\mathbf{37.99}$ & $\mathbf{34.85}$  \\
\hline
\end{tabular}
}
\caption{Final accuracy ($\%$) on the T200-B100-S5, T200-B100-S10, and T200-B100-S20 settings.}
\label{table:tinycontinuallearning}
\end{table}

\begin{table}
\centering
\resizebox{1\columnwidth}{!}{
\begin{tabular}{c|c|c|c|c}
\hline
Size of $\mathcal{B}$ &  Method &  C100-B0-S5 &  C100-B0-S10 &  C100-B0-S20 \\
\hline
\multirow{11}{*}{200} & ER~\cite{RiemerCALRTT19} & $21.94 \pm 0.83$ & $14.23 \pm 0.12$ & $9.90 \pm 1.67$  \\
& GEM~\cite{LopezPaz2017GradientEM} & $19.73 \pm 0.34$ & $13.20 \pm 0.21$ & $8.29 \pm 0.18$  \\
& AGEM~\cite{Chaudhry2019EfficientLL} & $17.97 \pm 0.26$ & $9.44 \pm 0.29$ & $4.88 \pm 0.09$  \\
& iCaRL~\cite{Rebuffi2017iCaRLIC} & $30.12 \pm 2.45$ & $22.38 \pm 2.79$  &  $12.62 \pm 1.43$ \\
& FDR~\cite{BenjaminRK19} & $22.84 \pm 1.49$ & $14.85 \pm 2.76$ & $6.70 \pm 0.79$  \\
& GSS~\cite{AljundiLGB19} & $19.44 \pm 2.83$ & $11.84 \pm 1.46$ & $6.42 \pm 1.24$ \\
& DER++~\cite{Buzzega2020DarkEF} & $27.46 \pm 1.16$ & $21.76 \pm 0.78$ & $15.16 \pm 1.53$ \\
& HAL~\cite{Chaudhry2021UsingHT} & $13.21 \pm 1.24$ & $9.67 \pm 1.67$ & $5.67 \pm 0.91$ \\
& ERT~\cite{Buzzega2021RethinkingER} & $21.61 \pm 0.87$ & $12.91 \pm 1.46$ & $10.14 \pm 1.96$ \\
& RM~\cite{Bang2021RainbowMC} & $32.23 \pm 1.09$ & $22.71 \pm 0.93$ & $15.15 \pm 2.14$  \\
& \bfseries Ours & $\mathbf{32.75 \pm 0.42}$ & $\mathbf{25.87 \pm 1.01}$ & $\mathbf{19.09 \pm 0.34}$ \\
\hline
\multirow{11}{*}{500} & ER~\cite{RiemerCALRTT19} & $27.97 \pm 0.33$ & $21.54 \pm 0.29$ & $15.36 \pm 1.15$ \\
& GEM~\cite{LopezPaz2017GradientEM} & $25.44 \pm 0.72$ & $18.48 \pm 1.34$ & $12.58 \pm 2.15$ \\
& AGEM~\cite{Chaudhry2019EfficientLL} & $18.75 \pm 0.51$ & $9.72 \pm 0.22$ & $5.97 \pm 1.13$ \\
& iCaRL~\cite{Rebuffi2017iCaRLIC} & $35.95 \pm 2.16$ & $30.25 \pm 1.86$ & $20.05 \pm 1.33$ \\
& FDR~\cite{BenjaminRK19} & $29.99 \pm 2.23$ & $22.81 \pm 2.81$ & $13.10 \pm 3.34$ \\
& GSS~\cite{AljundiLGB19} & $22.08 \pm 3.51$ & $13.72 \pm 2.64$ & $7.49 \pm 4.78$ \\
& DER++~\cite{Buzzega2020DarkEF} & $38.39 \pm 1.57$ & $36.15 \pm 1.10$ & $21.65 \pm 1.44$ \\
& HAL~\cite{Chaudhry2021UsingHT} & $16.74 \pm 3.51$ & $11.12 \pm 3.80$ & $9.71 \pm 2.91$ \\
& ERT~\cite{Buzzega2021RethinkingER} & $28.82 \pm 1.83$ & $23.00 \pm 0.58$ & $18.42 \pm 1.92$  \\
& RM~\cite{Bang2021RainbowMC} & $39.47 \pm 1.26$ & $32.52 \pm 1.53$ & $23.09 \pm 1.72$ \\
& \bfseries Ours & $\mathbf{40.85 \pm 0.91}$ & $\mathbf{33.55 \pm 0.53}$ & $\mathbf{28.78 \pm 0.87}$\\
\hline
\end{tabular}
}
\caption{Final accuracy ($\%$) on the C100-B0-S5, C100-B0-S10, and C100-B0-S20 settings.}
\label{table:secondgroup}
\end{table}

\begin{table}
\centering
\resizebox{1 \columnwidth}{!}{
\begin{tabular}{c|c c|c c|c c}
\hline
\multirow{2}{*}{Method} &  \multicolumn{2}{c|}{C100-B50-S5} &  \multicolumn{2}{c|}{C100-B50-S10} &  \multicolumn{2}{c}{C100-B40-S20}  \\
\cline{2-7}
& AIA$\uparrow$ ($\%$) & AF$\downarrow$ ($\%$) & AIA$\uparrow$ ($\%$) & AF$\downarrow$ ($\%$) & AIA$\uparrow$ ($\%$) & AF$\downarrow$ ($\%$)\\
\hline
iCaRLCNN~\cite{Rebuffi2017iCaRLIC} & $51.31$ & $42.13$ & $48.28$ & $45.69$ & $44.61$ & $43.54$\\
iCaRLNME~\cite{Rebuffi2017iCaRLIC} & $60.86$ & $24.90$ & $53.86$ & $28.32$ & $51.01$ & $35.53$\\
EEIL~\cite{CastroMGSA18} & $60.40$ & $23.36$ & $55.77$ & $26.65$ & $52.57$ & $32.40$\\
LUCIR~\cite{HouPLWL19} & $63.75$ & $21.00$ & $60.68$ & $25.12$ & $58.27$ & $28.65$\\
PASS~\cite{Zhu2021PrototypeAA} & $63.84$ & $25.20$ & $59.87$ & $30.25$ & $58.07$ & $30.61$\\
IL2A~\cite{Zhu2021ClassIncrementalLV} & $66.19$ & $29.57$ & $58.2$ & $39.75$ & $58.01$ & $48.66$ \\
HFA~\cite{HFAGAOZHI2022} & $66.96$ & $23.54$ & $63.7$ & $25.45$ & $59.02$ & $29.76$\\
\hline
\bfseries Ours & $\mathbf{67.44}$ & $\mathbf{21.94}$ & $\mathbf{65.93}$ & $\mathbf{23.76}$ & $\mathbf{61.85}$ & $\mathbf{24.71}$\\
\hline
\end{tabular}
}
\caption{Average incremental accuracy (denoted by 'AIA') and average forgetting (denoted by 'AF') on the C100-B50-S5, C100-B50-S10, and C100-B40-S20 settings.}
\label{table:averageaccuracyaverageforgetting}
\end{table}

\subsection{Main Results}

\begin{figure}[!t]
  \centering
  \includegraphics[width=0.42\textwidth]{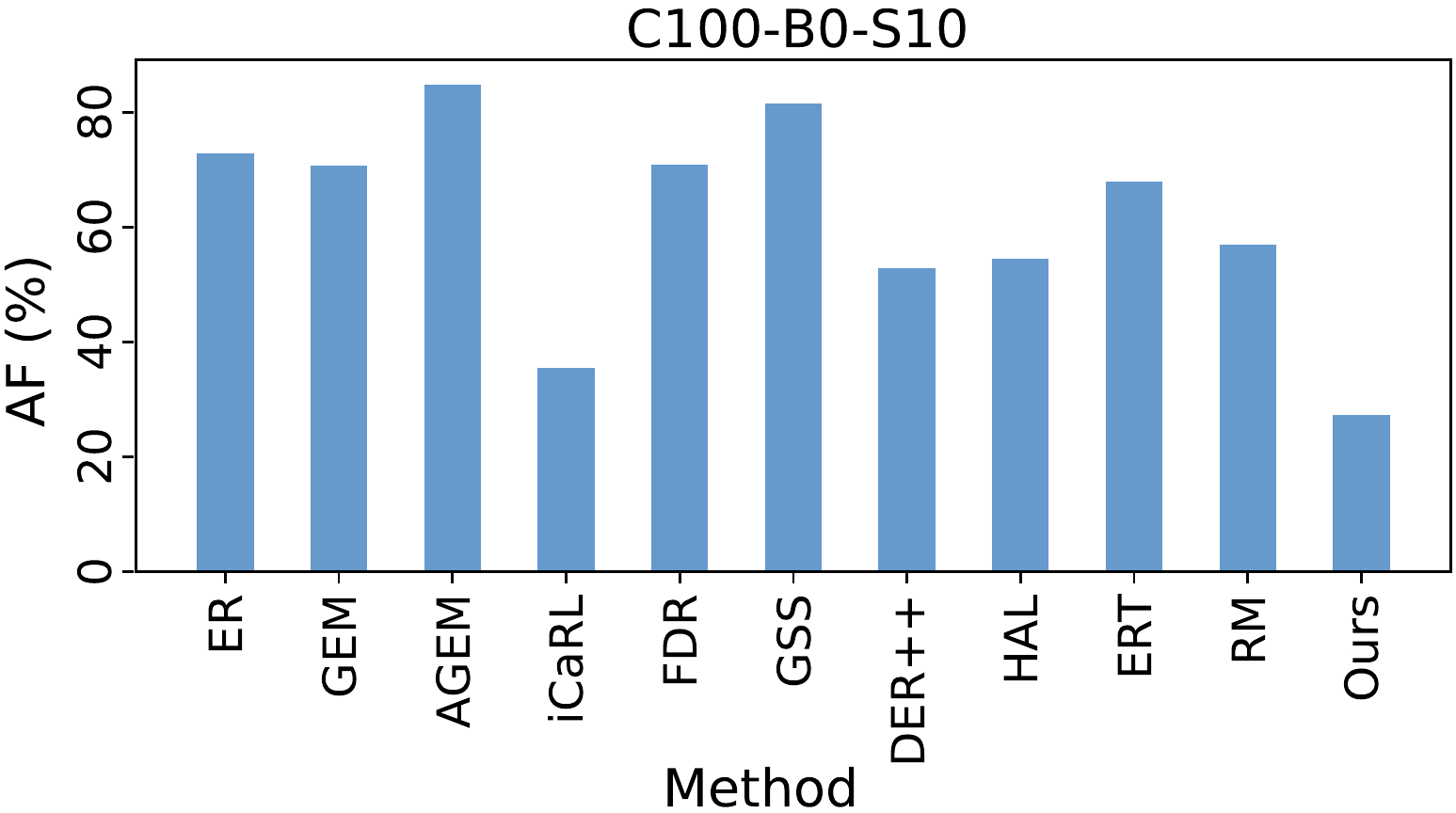}
  \caption{Average forgetting (AF, $\%$) on the C100-B0-S10 setting.}
  \label{fig:forgetting_hengban}
\end{figure}

\begin{figure}
    \centering
    \begin{subfigure}{0.155\textwidth}
      \centering
      \includegraphics[width=1\textwidth]{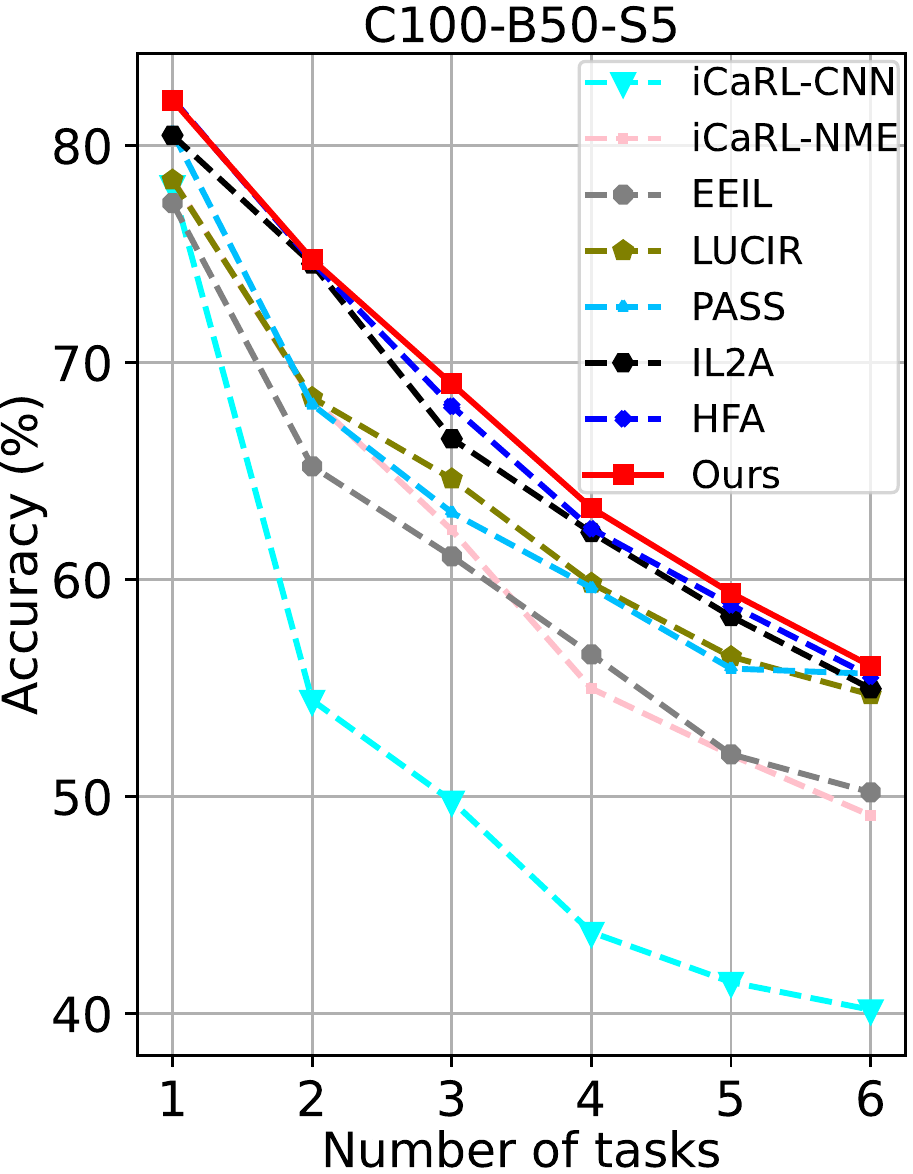}
    \end{subfigure}
    \begin{subfigure}{0.155\textwidth}
      \centering
       \includegraphics[width=1\textwidth]{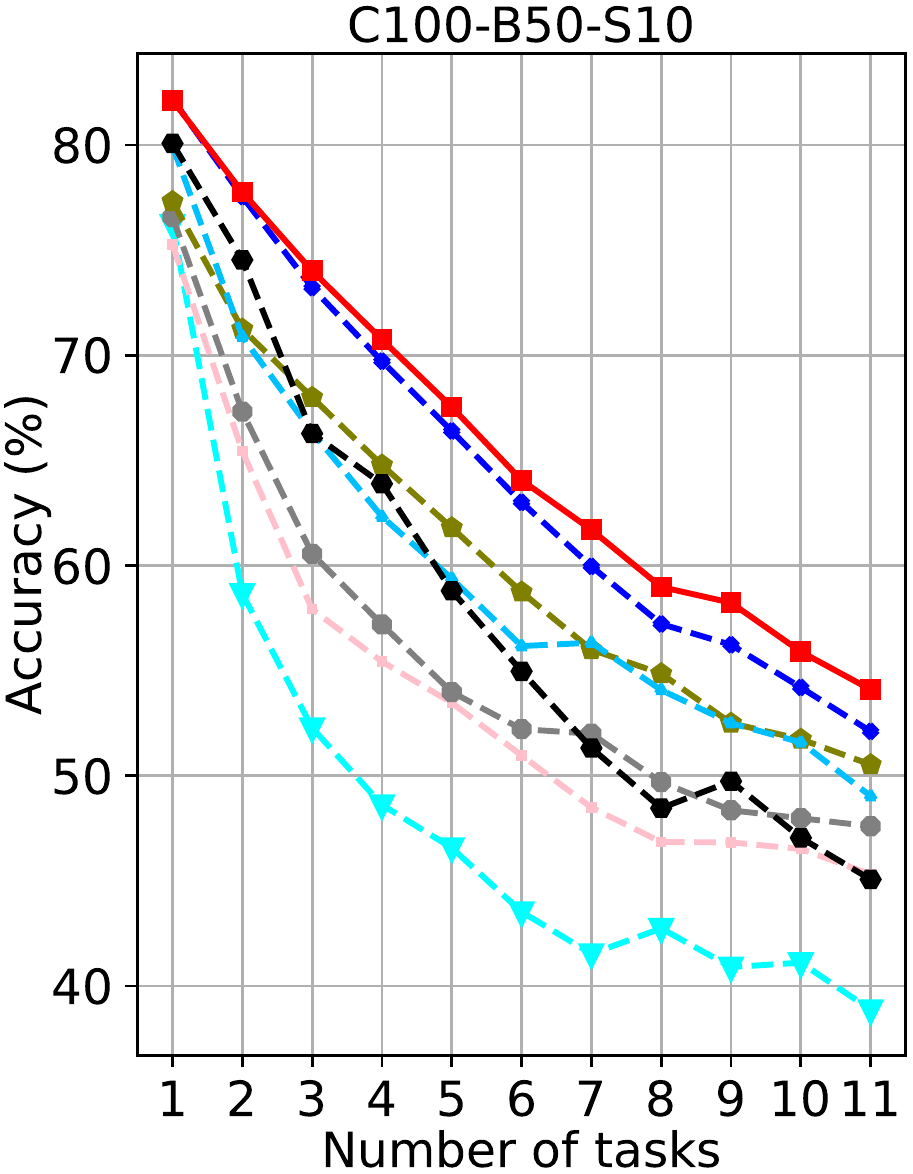}
    \end{subfigure} 
    \begin{subfigure}{0.155\textwidth}
      \centering
       \includegraphics[width=1\textwidth]{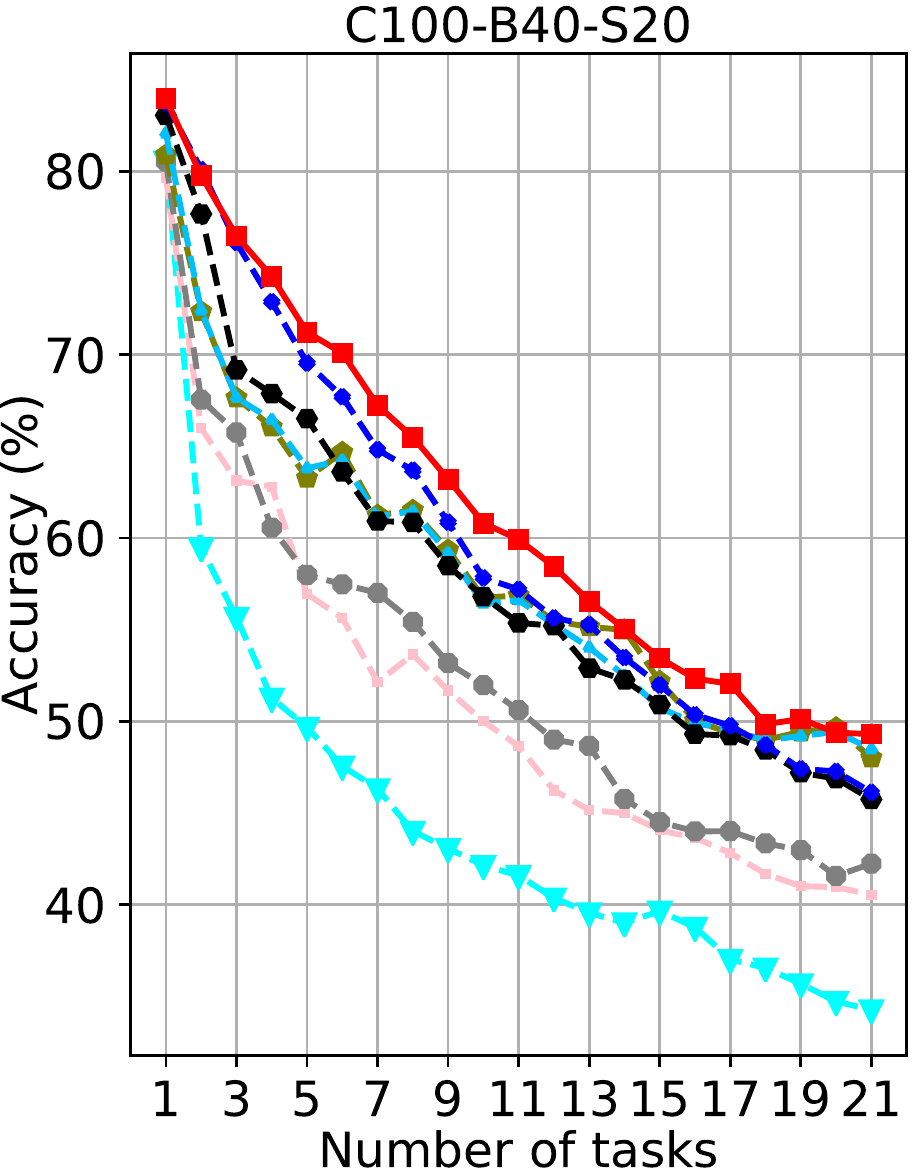}
    \end{subfigure}     
    \caption{Accuracy curves on the C100-B50-S5, C100-B50-S10, and C100-B40-S20 settings.}
\label{fig:accuracycurve}
\end{figure}

Final accuracies on the C100-B50-S5, C100-B50-S10, and C100-B40-S20 settings are shown in~\cref{table:cifarcontinuallearning}.
Final accuracies on the T200-B100-S5, T200-B100-S10, and T200-B100-S20 settings are shown in~\cref{table:tinycontinuallearning}.
Final accuracies on the C100-B0-S5, C100-B0-S10, and C100-B0-S20 settings are shown in~\cref{table:secondgroup}.
Average accuracies and average forgetting on the C100-B50-S5, C100-B50-S10, and C100-B40-S20 settings are shown in~\cref{table:averageaccuracyaverageforgetting}.
Average forgetting on the C100-B0-S20 setting is shown in~\cref{fig:forgetting_hengban}.
Accuracy curves on the C100-B50-S5, C100-B50-S10, and C100-B40-S20 settings are shown in~\cref{fig:accuracycurve}.
Our method achieves better performance than compared continual learning methods, having higher accuracies and lower forgetting. These results demonstrate the effectiveness of our method.

Take C100-B50-S10 as an example (in~\cref{table:cifarcontinuallearning}), our method finally achieves $54.12 \%$, $3.59 \%$ and $9.05 \%$ higher than the replay-based methods LUCIR~\cite{HouPLWL19} and IL2A~\cite{Zhu2021ClassIncrementalLV}, and $8.81 \%$ higher than the regularization-based method iCaRL~\cite{Rebuffi2017iCaRLIC}. This shows the effectiveness of exploring the geometric structure for continual learning.
Moreover, our performance is $2.01 \%$ higher than that of HFA~\cite{HFAGAOZHI2022} that uses a fixed hyperbolic geometry for the data stream. In contrast, our method expands the geometry of the underlying space to match the growing geometric structures of the data stream, leading to better performance.

In the second group of settings (in~\cref{table:secondgroup}), our method achieves good performance again on both the two sizes of the memory buffer.
With the steps in continual learning increasing, learning a good model becomes more and more challenging.
In this case, our improvements become larger and larger with the increase of steps compared with existing methods.
For example, when the size of the memory buffer is $200$, compared with RM~\cite{Bang2021RainbowMC}, our improvements are $0.52 \%$, $3.16\%$, and $3.94 \%$ on the C100-B0-S5, C100-B0-S10, C100-B0-S20 settings, respectively.
This shows that exploring date geometry benefits to continual learning, efficiently alleviating the catastrophic forgetting problem, especially in challenging settings.

\subsection{Ablation Study}

Our method includes three important components: the geometry incremental search scheme, the angular-regularization loss function, and the neighbor-robustness loss function. We conduct ablation experiments on the CIFAR-100 dataset to evaluate their effectiveness. 
We denote the geometry incremental search scheme, angular-regularization loss function, and the neighbor-robustness loss function as `GIS', `Global', and `Local', respectively. 
We first evaluate using the Euclidean geometry, that is, none of the three components is used, and the model is only trained by the cross-entropy loss in the Euclidean space using the memory buffer. 
Then, we add the geometry incremental search scheme to the model.
Next, we remove the geometry incremental search scheme and add the two proposed loss functions.
Finally, we evaluate the two loss functions by combining the geometry incremental search scheme with one of them. Results are shown in~\cref{table:ablation}.

We observe that all the three components have contributions to our improvement.
Comparing `GIS' with `Baseline' and `Ours' and  `Global+Local', the geometry incremental search scheme brings about $3 \%$ improvements on the C100-B50-S5 and C100-B40-S20 settings.
`Ours' has better performance than `GIS+Local', and `GIS+Global' has better performance than `GIS', showing the angular-regularization loss function helps solve the catastrophic forgetting problem in continual learning.
Similarly, the neighbor-robustness loss function also benefits to continual learning, especially in C100-B50-S10.
These results show that exploring data geometry via the designed components is an effective and promising way for continual learning.

\begin{table}
\centering
\resizebox{1\columnwidth}{!}{
\begin{tabular}{c|c c c}
\hline
 Method &  C100-B50-S5 &  C100-B50-S10 &  C100-B40-S20 \\
\hline
Euclidean & $50.63$ & $49.56$ & $43.60$ \\
GIS & $51.62$ & $49.86$ & $46.15$  \\
Global+Local & $52.63$ & $50.50$ & $44.45$\\
GIS+Global & $52.49$ & $50.33$ & $47.31$  \\
GIS+Local & $53.94$ & $53.40$ & $46.84$ \\
\hline
\bfseries Ours & $\mathbf{56.03}$ & $\mathbf{54.31}$ & $\mathbf{49.32}$  \\
\hline
\end{tabular}
}
\caption{Final accuracy ($\%$) of ablation results on the C100-B50-S5, C100-B50-S10, and C100-B40-S20 settings.}
\label{table:ablation}
\end{table}

\section{Conclusion}
In this paper, we have presented that exploring data geometry benefits to continual learning, through which the forgetting issue of old data can be prevented by preserving geometric structures of data and learning from new data benefits from expanding the geometry of the underlying space.
The introduced geometry incremental search scheme can produce suitable geometry for new data, avoiding data distortions.
The proposed angular-regularization loss function can well preserve global structures of old data, and the neighbor-robustness loss can make local structures more discriminative.
Experiments on multiple settings show that our method achieves better performance than continual learning methods designed in Euclidean space.
In the current method, the submanifold pool has a finite number of CCSs, which may limit the ability of the mixed-curvature space. In the future, we are going to study more flexible schemes, where the submanifold pool has an infinite number of CCSs with arbitrary dimensions.

\noindent \textbf{Acknowledgements.}
This work was supported by the Natural Science Foundation of China (NSFC)
under Grants No. 62172041 and No. 62176021, and Shenzhen National Science Foundation (Shenzhen Stable Support Fund for College Researchers), No. 20200829143245001.

{\small
\bibliographystyle{ieee_fullname}
\bibliography{egbib}

\begin{thebibliography}{10}\itemsep=-1pt

\bibitem{AljundiLGB19}
Rahaf Aljundi, Min Lin, Baptiste Goujaud, and Yoshua Bengio.
\newblock Gradient based sample selection for online continual learning.
\newblock In {\em Advances in Neural Information Processing Systems (NeurIPS)},
  pages 11816--11825, 2019.

\bibitem{AtighKM21}
Mina~Ghadimi Atigh, Martin Keller{-}Ressel, and Pascal Mettes.
\newblock Hyperbolic busemann learning with ideal prototypes.
\newblock In {\em Advances in Neural Information Processing Systems (NeurIPS)},
  pages 103--115, 2021.

\bibitem{Bachmann2019ConstantCG}
G. Bachmann, Gary B{\'e}cigneul, and Octavian-Eugen Ganea.
\newblock Constant curvature graph convolutional networks.
\newblock In {\em International Conference on Machine Learning (ICML)}, pages
  486--496, 2020.

\bibitem{Bang2021RainbowMC}
Jihwan Bang, Heesu Kim, Young~Joon Yoo, Jung-Woo Ha, and Jonghyun Choi.
\newblock Rainbow memory: Continual learning with a memory of diverse samples.
\newblock {\em IEEE/CVF Conference on Computer Vision and Pattern Recognition
  (CVPR)}, pages 8214--8223, 2021.

\bibitem{BenjaminRK19}
Ari~S. Benjamin, David Rolnick, and Konrad~P. K{\"{o}}rding.
\newblock Measuring and regularizing networks in function space.
\newblock In {\em International Conference on Learning Representations (ICLR)},
  2019.

\bibitem{BronsteinBLSV17}
Michael~M. Bronstein, Joan Bruna, Yann LeCun, Arthur Szlam, and Pierre
  Vandergheynst.
\newblock Geometric deep learning: Going beyond euclidean data.
\newblock {\em {IEEE} Signal Process. Mag.}, 34(4):18--42, 2017.

\bibitem{buss2001spherical}
Samuel~R Buss and Jay~P Fillmore.
\newblock Spherical averages and applications to spherical splines and
  interpolation.
\newblock {\em ACM Transactions on Graphics (TOG)}, 20(2):95--126, 2001.

\bibitem{Buzzega2020DarkEF}
Pietro Buzzega, Matteo Boschini, Angelo Porrello, Davide Abati, and Simone
  Calderara.
\newblock Dark experience for general continual learning: a strong, simple
  baseline.
\newblock In {\em Advances in Neural Information Processing Systems (NeurIPS)},
  pages 15920--15930, 2020.

\bibitem{Buzzega2021RethinkingER}
Pietro Buzzega, Matteo Boschini, Angelo Porrello, and Simone Calderara.
\newblock Rethinking experience replay: a bag of tricks for continual learning.
\newblock {\em International Conference on Pattern Recognition (ICPR)}, pages
  2180--2187, 2021.

\bibitem{cannon1997hyperbolic}
James~W Cannon, William~J Floyd, Richard Kenyon, Walter~R Parry, et~al.
\newblock Hyperbolic geometry.
\newblock {\em Flavors of geometry}, 31(59-115):2, 1997.

\bibitem{CastroMGSA18}
Francisco~M. Castro, Manuel~J. Mar{\'{\i}}n{-}Jim{\'{e}}nez, Nicol{\'{a}}s
  Guil, Cordelia Schmid, and Karteek Alahari.
\newblock End-to-end incremental learning.
\newblock In {\em European Conference on Computer Vision (ECCV)}, pages
  241--257, 2018.

\bibitem{Chaudhry2021UsingHT}
Arslan Chaudhry, Albert Gordo, Puneet~Kumar Dokania, Philip H.~S. Torr, and
  David Lopez-Paz.
\newblock Using hindsight to anchor past knowledge in continual learning.
\newblock In {\em AAAI Conference on Artificial Intelligence (AAAI)}, 2021.

\bibitem{Chaudhry2019EfficientLL}
Arslan Chaudhry, Marc'Aurelio Ranzato, Marcus Rohrbach, and Mohamed Elhoseiny.
\newblock Efficient lifelong learning with a-gem.
\newblock In {\em International Conference on Learning Representations (ICLR)},
  2019.

\bibitem{Cheraghian2021SemanticawareKD}
Ali Cheraghian, Shafin Rahman, Pengfei Fang, Soumava~Kumar Roy, Lars Petersson,
  and Mehrtash~Tafazzoli Harandi.
\newblock Semantic-aware knowledge distillation for few-shot class-incremental
  learning.
\newblock In {\em IEEE/CVF Conference on Computer Vision and Pattern
  Recognition (CVPR)}, pages 2534--2543, 2021.

\bibitem{FangHP21}
Pengfei Fang, Mehrtash Harandi, and Lars Petersson.
\newblock Kernel methods in hyperbolic spaces.
\newblock In {\em IEEE/CVF International Conference on Computer Vision (ICCV)},
  pages 10645--10654, 2021.

\bibitem{Gao2022CurvatureAdaptiveMF}
Zhi Gao, Yuwei Wu, Mehrtash Harandi, and Yunde Jia.
\newblock Curvature-adaptive meta-learning for fast adaptation to manifold
  data.
\newblock {\em IEEE Transactions on Pattern Analysis and Machine Intelligence
  (T-PAMI)}, 45(2):1545--1562, 2023.

\bibitem{GaoWJH21}
Zhi Gao, Yuwei Wu, Yunde Jia, and Mehrtash Harandi.
\newblock Curvature generation in curved spaces for few-shot learning.
\newblock In {\em IEEE/CVF International Conference on Computer Vision (ICCV)},
  pages 8671--8680, 2021.

\bibitem{HFAGAOZHI2022}
Zhi Gao, Yuwei Wu, Yunde Jia, and Mehrtash Harandi.
\newblock Hyperbolic feature augmentation via distribution estimation and
  infinite sampling on manifolds.
\newblock In {\em Advances in Neural Information Processing Systems (NeurIPS)},
  2022.

\bibitem{gu2018learning}
Albert Gu, Frederic Sala, Beliz Gunel, and Christopher R{\'e}.
\newblock Learning mixed-curvature representations in product spaces.
\newblock In {\em International Conference on Learning Representations (ICLR)},
  2019.

\bibitem{He2016DeepRL}
Kaiming He, X. Zhang, Shaoqing Ren, and Jian Sun.
\newblock Deep residual learning for image recognition.
\newblock {\em IEEE Conference on Computer Vision and Pattern Recognition
  (CVPR)}, pages 770--778, 2016.

\bibitem{HouPLWL19}
Saihui Hou, Xinyu Pan, Chen~Change Loy, Zilei Wang, and Dahua Lin.
\newblock Learning a unified classifier incrementally via rebalancing.
\newblock In {\em IEEE/CVF Conference on Computer Vision and Pattern
  Recognition (CVPR)}, pages 831--839, 2019.

\bibitem{khrulkov2020hyperbolic}
Valentin Khrulkov, Leyla Mirvakhabova, Evgeniya Ustinova, Ivan Oseledets, and
  Victor Lempitsky.
\newblock Hyperbolic image embeddings.
\newblock In {\em IEEE/CVF Conference on Computer Vision and Pattern
  Recognition (CVPR)}, pages 6418--6428, 2020.

\bibitem{Kirkpatrick2017OvercomingCF}
James Kirkpatrick, Razvan Pascanu, Neil~C. Rabinowitz, Joel Veness, Guillaume
  Desjardins, Andrei~A. Rusu, Kieran Milan, John Quan, Tiago Ramalho, Agnieszka
  Grabska-Barwinska, Demis Hassabis, Claudia Clopath, Dharshan Kumaran, and
  Raia Hadsell.
\newblock Overcoming catastrophic forgetting in neural networks.
\newblock {\em National Academy of Sciences}, 114:3521 -- 3526, 2017.

\bibitem{Kong2022BalancingSA}
Yajing Kong, Liu Liu, Zhen Wang, and Dacheng Tao.
\newblock Balancing stability and plasticity through advanced null space in
  continual learning.
\newblock In {\em European Conference on Computer Vision (ECCV)}, pages
  219--236, 2022.

\bibitem{Krizhevsky2009LearningML}
Alex Krizhevsky.
\newblock Learning multiple layers of features from tiny images.
\newblock 2009.

\bibitem{Le2015TinyIV}
Ya Le and Xuan~S. Yang.
\newblock Tiny imagenet visual recognition challenge.
\newblock 2015.

\bibitem{Li_2021_CVPR}
Shen Li, Jianqing Xu, Xiaqing Xu, Pengcheng Shen, Shaoxin Li, and Bryan Hooi.
\newblock Spherical confidence learning for face recognition.
\newblock In {\em IEEE/CVF Conference on Computer Vision and Pattern
  Recognition (CVPR)}, pages 15629--15637, 2021.

\bibitem{Li2018LearningWF}
Zhizhong Li and Derek Hoiem.
\newblock Learning without forgetting.
\newblock {\em IEEE Transactions on Pattern Analysis and Machine Intelligence
  (T-PAMI)}, 40:2935--2947, 2018.

\bibitem{long2020searching}
Teng Long, Pascal Mettes, Heng~Tao Shen, and Cees~GM Snoek.
\newblock Searching for actions on the hyperbole.
\newblock In {\em IEEE/CVF Conference on Computer Vision and Pattern
  Recognition (CVPR)}, pages 1141--1150, 2020.

\bibitem{LopezPaz2017GradientEM}
David Lopez-Paz and Marc'Aurelio Ranzato.
\newblock Gradient episodic memory for continual learning.
\newblock In {\em Advances in Neural Information Processing Systems (NeurIPS)},
  pages 6467--6476, 2017.

\bibitem{lou2021learning}
Aaron Lou, Maximilian Nickel, Mustafa Mukadam, and Brandon Amos.
\newblock Learning complex geometric structures from data with deep riemannian
  manifolds.
\newblock 2021.

\bibitem{Mallya2018PackNetAM}
Arun Mallya and Svetlana Lazebnik.
\newblock Packnet: Adding multiple tasks to a single network by iterative
  pruning.
\newblock In {\em IEEE/CVF Conference on Computer Vision and Pattern
  Recognition (CVPR)}, pages 7765--7773, 2018.

\bibitem{nickel2017poincare}
Maximillian Nickel and Douwe Kiela.
\newblock Poincar{\'e} embeddings for learning hierarchical representations.
\newblock {\em Advances in neural information processing systems (NeurIPS)},
  2017.

\bibitem{ParkKLC19}
Wonpyo Park, Dongju Kim, Yan Lu, and Minsu Cho.
\newblock Relational knowledge distillation.
\newblock In {\em IEEE/CVF Conference on Computer Vision and Pattern
  Recognition (CVPR)}, pages 3967--3976, 2019.

\bibitem{QiYLL21}
Guodong Qi, Huimin Yu, Zhaohui Lu, and Shuzhao Li.
\newblock Transductive few-shot classification on the oblique manifold.
\newblock In {\em IEEE/CVF International Conference on Computer Vision (ICCV)},
  pages 8392--8402, 2021.

\bibitem{Rebuffi2017iCaRLIC}
Sylvestre-Alvise Rebuffi, Alexander Kolesnikov, G. Sperl, and Christoph~H.
  Lampert.
\newblock icarl: Incremental classifier and representation learning.
\newblock In {\em IEEE/CVF Conference on Computer Vision and Pattern
  Recognition (CVPR)}, pages 5533--5542, 2017.

\bibitem{RiemerCALRTT19}
Matthew Riemer, Ignacio Cases, Robert Ajemian, Miao Liu, Irina Rish, Yuhai Tu,
  and Gerald Tesauro.
\newblock Learning to learn without forgetting by maximizing transfer and
  minimizing interference.
\newblock In {\em International Conference on Learning Representations (ICLR)},
  2019.

\bibitem{Rusu2016ProgressiveNN}
Andrei~A. Rusu, Neil~C. Rabinowitz, Guillaume Desjardins, Hubert Soyer, James
  Kirkpatrick, Koray Kavukcuoglu, Razvan Pascanu, and Raia Hadsell.
\newblock Progressive neural networks.
\newblock {\em ArXiv}, abs/1606.04671, 2016.

\bibitem{Serr2018OvercomingCF}
Joan Serr{\`a}, D{\'i}dac Sur{\'i}s, Marius Miron, and Alexandros Karatzoglou.
\newblock Overcoming catastrophic forgetting with hard attention to the task.
\newblock In {\em International Conference on Machine Learning (ICML)}, 2018.

\bibitem{Shevkunov2021OverlappingSF}
Kirill Shevkunov and Liudmila Prokhorenkova.
\newblock Overlapping spaces for compact graph representations.
\newblock In {\em Advances in Neural Information Processing Systems (NeurIPS)},
  pages 11665--11677, 2021.

\bibitem{Shin2017ContinualLW}
Hanul Shin, Jung~Kwon Lee, Jaehong Kim, and Jiwon Kim.
\newblock Continual learning with deep generative replay.
\newblock In {\em Advances in Neural Information Processing Systems (NeurIPS)},
  pages 2990--2999, 2017.

\bibitem{skopek2019mixed}
Ondrej Skopek, Octavian-Eugen Ganea, and Gary B{\'e}cigneul.
\newblock Mixed-curvature variational autoencoders.
\newblock In {\em International Conference on Learning Representations (ICLR)},
  2020.

\bibitem{Sun2022ASM}
Li Sun, Zhongbao Zhang, Junda Ye, Hao Peng, Jiawei Zhang, Sen Su, and Philip~S.
  Yu.
\newblock A self-supervised mixed-curvature graph neural network.
\newblock In {\em AAAI Conference on Artificial Intelligence (AAAI)}, 2022.

\bibitem{Tao2020BiObjectiveCL}
Xiaoyu Tao, Xiaopeng Hong, Xinyuan Chang, and Yihong Gong.
\newblock Bi-objective continual learning: Learning 'new' while consolidating
  'known'.
\newblock In {\em AAAI Conference on Artificial Intelligence (AAAI)}, pages
  5989--5996, 2020.

\bibitem{Wang2021TrainingNI}
Shipeng Wang, Xiaorong Li, Jian Sun, and Zongben Xu.
\newblock Training networks in null space of feature covariance for continual
  learning.
\newblock In {\em IEEE/CVF Conference on Computer Vision and Pattern
  Recognition (CVPR)}, pages 184--193, 2021.

\bibitem{WangWSWNAXYC21}
Shen Wang, Xiaokai Wei, C{\'{\i}}cero~Nogueira dos Santos, Zhiguo Wang, Ramesh
  Nallapati, Andrew~O. Arnold, Bing Xiang, Philip~S. Yu, and Isabel~F. Cruz.
\newblock Mixed-curvature multi-relational graph neural network for knowledge
  graph completion.
\newblock In {\em International Conference of World Wide Web (WWW)}, pages
  1761--1771, 2021.

\bibitem{Wang2022ContinualLW}
Zhen Wang, Liu Liu, Yiqun Duan, Yajing Kong, and Dacheng Tao.
\newblock Continual learning with lifelong vision transformer.
\newblock In {\em IEEE/CVF Conference on Computer Vision and Pattern
  Recognition (CVPR)}, pages 171--181, 2022.

\bibitem{XuWWLWYDZZXZ22}
Zhirong Xu, Shiyang Wen, Junshan Wang, Guojun Liu, Liang Wang, Zhi Yang, Lei
  Ding, Yan Zhang, Di Zhang, Jian Xu, and Bo Zheng.
\newblock {AMCAD:} adaptive mixed-curvature representation based advertisement
  retrieval system.
\newblock In {\em IEEE International Conference on Data Engineering (ICDE)},
  pages 3439--3452, 2022.

\bibitem{Yan2021DERDE}
Shipeng Yan, Jiangwei Xie, and Xuming He.
\newblock Der: Dynamically expandable representation for class incremental
  learning.
\newblock In {\em IEEE/CVF Conference on Computer Vision and Pattern
  Recognition (CVPR)}, pages 3013--3022, 2021.

\bibitem{Yu2020SemanticDC}
Lu Yu, Bartlomiej Twardowski, Xialei Liu, Luis Herranz, Kai Wang, Yongmei
  Cheng, Shangling Jui, and Joost van~de Weijer.
\newblock Semantic drift compensation for class-incremental learning.
\newblock In {\em IEEE/CVF Conference on Computer Vision and Pattern
  Recognition (CVPR)}, pages 6980--6989, 2020.

\bibitem{Zenke2017ContinualLT}
Friedemann Zenke, Ben Poole, and Surya Ganguli.
\newblock Continual learning through synaptic intelligence.
\newblock In {\em International Conference on Machine Learning (ICML)}, pages
  3987--3995, 2017.

\bibitem{zhang2021switch}
Shuai Zhang, Yi Tay, Wenqi Jiang, Da-cheng Juan, and Ce Zhang.
\newblock Switch spaces: Learning product spaces with sparse gating.
\newblock {\em arXiv preprint arXiv:2102.08688}, 2021.

\bibitem{Zhu2021ClassIncrementalLV}
Fei Zhu, Zhen Cheng, Xu-Yao Zhang, and Cheng-Lin Liu.
\newblock Class-incremental learning via dual augmentation.
\newblock In {\em Advances in Neural Information Processing Systems (NeurIPS)},
  pages 14306--14318, 2021.

\bibitem{Zhu2021PrototypeAA}
Fei Zhu, Xu-Yao Zhang, Chuan Wang, Fei Yin, and Cheng-Lin Liu.
\newblock Prototype augmentation and self-supervision for incremental learning.
\newblock In {\em IEEE/CVF Conference on Computer Vision and Pattern
  Recognition (CVPR)}, pages 5867--5876, 2021.

\end{thebibliography}
}

\end{document}